\documentclass[letterpaper]{article} 
\usepackage[]{aaai24}  
\usepackage{times}  
\usepackage{helvet}  
\usepackage{courier}  
\usepackage[hyphens]{url}  
\usepackage{graphicx} 
\usepackage{natbib}  
\usepackage{caption} 
\usepackage{algorithm}
\usepackage{algorithmic}
\usepackage{amssymb}
\usepackage{amsmath}
\usepackage{xcolor}
\usepackage{float}
\usepackage{newfloat}
\usepackage{listings}
\DeclareCaptionStyle{ruled}{labelfont=normalfont,labelsep=colon,strut=off} 
\floatstyle{ruled}
\newfloat{listing}{tb}{lst}{}
\floatname{listing}{Listing}
\title{HI-GAN: Hierarchical Inpainting GAN with Auxiliary Inputs for Combined RGB and Depth Inpainting}
\usepackage{bibentry}
\begin{document}
%

\author{
    Ankan Dash,
    Jingyi Gu,
    Guiling Wang
}
\affiliations{
  New Jersey Institute of Technology, Newark, USA\\
    \{ad892, jg95, gwang\}@njit.edu
}
\maketitle
\begin{abstract}
Inpainting involves filling in missing pixels or areas in an image, a crucial technique employed in Mixed Reality environments for various applications, particularly in Diminished Reality (DR) where content is removed from a user's visual environment. Existing methods rely on digital replacement techniques which necessitate multiple cameras and incur high costs. AR devices and smartphones use ToF depth sensors to capture scene depth maps aligned with RGB images. Despite speed and affordability, ToF cameras create imperfect depth maps with missing pixels. 
To address the above challenges, we propose Hierarchical Inpainting GAN (\emph{HI-GAN}), a novel approach comprising three GANs in a hierarchical fashion for RGBD inpainting. 
\emph{EdgeGAN} and \emph{LabelGAN} inpaint masked edge and segmentation label images respectively, while \emph{CombinedRGBD-GAN} combines their latent representation outputs and performs RGB and Depth inpainting. Edge images and particularly segmentation label images as auxiliary inputs significantly enhance inpainting performance by complementary context and hierarchical optimization. We believe we make the first attempt to incorporate label images into inpainting process.
Unlike previous approaches requiring multiple sequential models and separate outputs, our work operates in an end-to-end manner, training all three models simultaneously and hierarchically. Specifically, \emph{EdgeGAN} and \emph{LabelGAN} are first optimized separately and further optimized inside \emph{CombinedRGBD-GAN} to enhance inpainting quality. Experiments demonstrate that \emph{HI-GAN} works seamlessly and achieves overall superior performance compared with existing approaches. 
\end{abstract}

\section{Introduction}
Inpainting restores missing pixels in images, aiding image restoration when content is incomplete or undesirable. Users can make holes, i.e., a section of missing pixels, where specific objects used to be and fill them with appropriate information to remove them from the image. 
One such example is to remove the watermarks on an image using inpainting. 

Inpainting of missing pixels is essential in many tasks. 
(1) In Augmented Reality (AR), blending digital content seamlessly with the physical world is vital.  For this reason, smartphones and augmented reality headsets are fitted with depth sensors that record the distance between the device and locations in the actual world as a depth map corresponding to the scene's pixels. A depth image's information can be utilized to precisely place virtual objects in front of or behind real-world objects, creating immersive and realistic user experiences. 
Time of Flight or ToF sensors are widely used to capture depth maps because they are cheap and in fact can be found in consumer smartphones such as the Samsung Galaxy A80. Unfortunately, there are issues with ToF cameras that degrade the quality of the depth data they capture and cause the output to have missing pixels \cite{10.1145/3517260}.   
(2) A component of augmented reality known as Diminished Reality, or DR, is the perception of an object being removed from its surroundings. For the successful removal of an object, it is vital to fill in the missing pixels resulting from the object removal with plausible pixels. To create DR, digital content that emulates the area behind an object is displayed after the object has been removed. The user thus believes the object is no longer present. 

The majority of today's state-of-the-art image inpainting approaches focus solely on RGB images rather than RGBD ones, and the remaining techniques concentrate solely on inpainting Depth images individually rather than RGB and Depth together. 
Note that an RGB-D image offers pixel-level depth information matched to the relevant image pixels. Understanding the geometric relationships in a scene requires accurate depth estimation. Depth images have diverse uses in 3D computer graphics and computer vision. Depth maps simulate shallow depth of field, blurring elements for a focus effect. Another application is detecting and excluding hidden objects from rendering. Depth maps are also vital for modeling and reconstructing 3D shapes. \cite{3DVAE}. Lee et al.\cite{lee2011depth} used depth maps along with RGB images for real-time 3D object detection for augmented reality. Their approach demonstrated the advantages of using depth maps and RGB images by detecting objects with complex structures, robust detection under varying lighting conditions, etc. Thus, depth images are necessary for inpainting task.

While certain existing works have incorporated auxiliary inputs to enhance inpainting process, they still suffer from limitations. EdgeConnect \cite{Nazeri_2019_ICCV} with edge images for inpainting of only RGB images does not operate in an end-to-end manner. It demands sequential training of multiple models, where the output of the previous model serves as the input to the next one during training. Boundary Aware \cite{boundaryAware} leverages depth images to guide RGB inpainting. However, it relies on generating depth data through monocular depth estimation, introducing potential discrepancies between predicted and actual ground truth depth images. To solve this issue, we aim to employ authentic depth images that provide accurate depth information, along with RGB images which are prevalent in mixed reality environments. 

As the era of Deep Learning exploded, the researchers shifted towards Neural Networks based Inpainting methods \cite{Nazeri_2019_ICCV,gatedConv,PartialConv, boundaryAware, 8100211, 10.1145/3343031.3351002}. the prevailing focus within this realm has been either on RGB image inpainting\cite{Nazeri_2019_ICCV,gatedConv,PartialConv, boundaryAware} or Depth inpainting\cite{Zhang_2018_CVPR,huang2019indoor,inpaint_transformer, 10.1145/3517260} in isolation.
Given the pivotal role of depth information in AR/VR applications, therefore,  
we aim to perform inpainting across both RGB and depth modalities. We leverage Deep Learning techniques to solve the missing pixel problem not only for RGB images but also extend to RGBD images. 

To fill the above gaps, we propose Hierarchical Inpainting GAN (\emph{HI-GAN}), an innovative framework with auxiliary inputs and their corresponding regularizers to augment RGBD inpainting. Specifically, given masked RGB and depth images, along with auxiliary inputs including masked edge images and masked semantic segmentation labels, \emph{HI-GAN} aims to reconstruct both RGB and depth images without missing pixels. 
The integration of edge and semantic segmentation label images guides the inpainting from two key aspects: complementary context and hierarchical optimization. Contextually, edge images provide valuable information on object boundaries; label images facilitate object-aware inpainting by supplying information regarding not only boundaries and shapes but also specific objects and classes, enriching the inpainting process and enhancing precision. 
Optimization-wise, three GANs collaborate together during the training process: \emph{EdgeGAN} for edge inpainting, \emph{LabelGAN} for semantic segmentation label inpainting, and \emph{CombinedRGBD-GAN} for RGB and Depth inpainting. 
The former two models serve as regularizers and generate latent edge and label representations, which are subsequently fused into \emph{CombinedRGBD-GAN}. This design facilitates Hierarchical Optimization, where \emph{EdgeGAN} and \emph{LabelGAN} are further optimized inside the \emph{CombinedRGBD-GAN} beyond their individual adversarial feedback training. 

Our contributions are listed as follows:
\begin{itemize}
    \item We propose \textbf{Hierarchical Inpainting GAN (\emph{HI-GAN})}, a novel framework consisting of three GANs operating in an end-to-end fashion for RGBD inpainting. We make the first attempt to incorporate segmentation label images as auxiliary inputs to augment RGBD inpainting.
    \item Auxiliary inputs including edge and label images guide the RGBD inpainting process in terms of context and optimization. Contextually, they offer crucial insights into boundaries and specific objects \& classes respectively. During optimization,  \emph{EdgeGAN} and \emph{LabelGAN} serves as regularizers, trained separately and further optimized inside \emph{CombinedRGBD-GAN}.
    \item Extensive experiments validate that edge images and especially semantic segmentation label images along with our novel training strategy vastly improve RGBD inpainting performance, surpassing state-of-the-art methods.
\end{itemize}

\section{Literature Review}
DR inpainting techniques have been utilized to capture the missing pixels using multiple view observations, which often necessitates real-time observations from multiple cameras. 3D PixMix \cite{8699212} integrated both RGB and Depth information in the inpainting process, to illustrate a similar methodology. However, multi-view techniques face limitations when the target's background is obscured in alternate views. Other approaches rely on inpainting techniques that do not hinge on multiple view observations, instead replacing missing pixels using available image data. 

Early efforts on deep learning based methods inpaint missing pixels for RGB and Depth images separately.
Partial Convolutions \cite{PartialConv} was proposed for RGB inpainting of irregular holes. Their partial convolution layer comprises masked and re-normalized convolution operations followed by a mask-update setup.
Huang et al.\cite{huang2019indoor} performed Depth inpainting with self-attention mechanism and boundary consistency. 
Zhang et al.\cite{Zhang_2018_CVPR} proposed a deep learning framework for Depth channel inpainting of an RGB-D image, employing the information from aligned RGB images. It uses 4 major steps, involving Estimating Surface normal, boundary detection, occlusion weight, and global optimization. 
Fujii et al.\cite{FujiiRGBD} introduced a GAN based approach with late fusion for RGBD image inpainting, yet it exclusively caters to images with missing pixels in regular rectangle-shaped areas. This is unrealistic as real-world missing pixels are generally randomly distributed and lack uniform shapes.

To enhance the performance of RGB inpainting, efforts have been directed towards incorporating  auxialiary inputs, such as depth images, edge images, and semantic segmentation labels. For instance, EdgeConnect \cite{Nazeri_2019_ICCV} consists of two stages, an edge generator and an image completion network. In the first stage, the edge generator recovers the missing region in the edge images. In the second stage, the image completion network utilizes the generated edges as a guide to fill in the missing regions of the RGB image. A notable limitation is its non-end-to-end training process, requiring separate training phases and a specific model training sequence to achieve desired results.  Boundary Aware \cite{boundaryAware} integrated edge images alongside depth images generated through monocular depth estimation to improve RGB inpainting. However, the depth images are generated by monocular depth estimation and not always be as informative as real depth images. 

The existing landscape reveals a predominant focus on separate RGB and Depth inpainting, instead of RGBD inpainting. Certain approaches exhibit drawbacks in training strategies, ranging from non-end-to-end training to the necessity of specific model training sequences. 
In mixed reality environments, both rgb and depth images are crucial. Therefore, we present a innovative framework, which takes masked RGB, depth, edge, segmentation label as input to seamlessly fill in missing pixels. Our approach demonstrates superior performance compared to various state-of-the-art baseline models, all achieved without resorting to complex attention mechanisms or intricate training strategies.

\begin{figure*}[h!]
    \centering
    \includegraphics[width=1.0\textwidth]{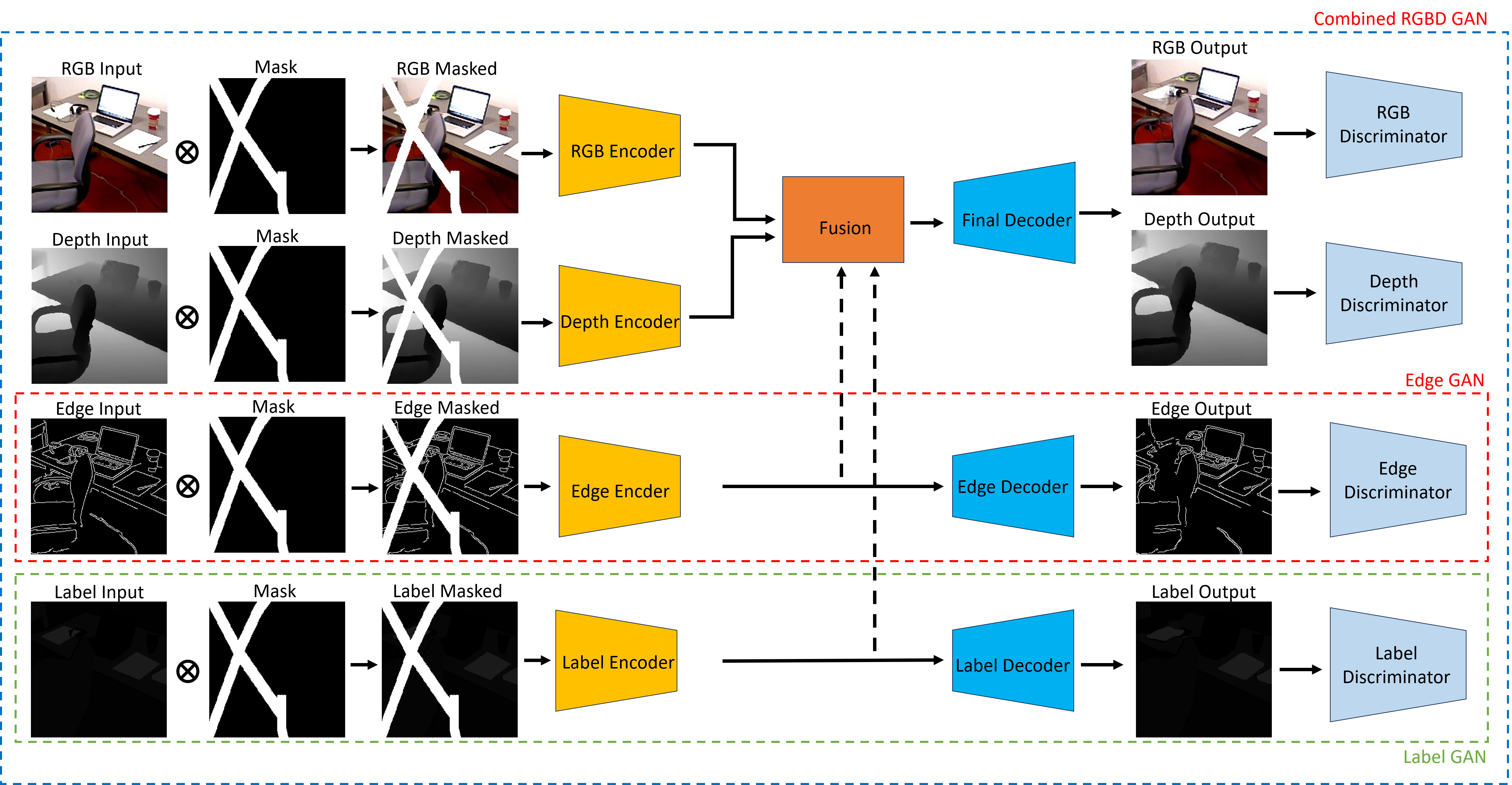}
    \caption{The architecture of \emph{HI-GAN} for RGBD inpainting with auxiliary inputs. The mask images are applied to the input RGB, Depth, Edge, and Label images. The Fusion module concatenates all latent representations from individual encoders.}
    \label{fig:hGAN}
\end{figure*}

\section{Preliminary}
\subsection{Problem Formulation}
In this paper, we propose \emph{HI-GAN} for simultaneous RGB and Depth image inpainting.
The dataset consists $N$ samples of images
$(\mathbf{y}_{r},\mathbf{y}_{d},\mathbf{y}_{e},\mathbf{y}_{l})$, . Each sample contains a 3-channel RGB image $\mathbf{y}_{r}$, a depth image $\mathbf{y}_{d}$, an edge image $\mathbf{y}_{e}$, and a label image $\mathbf{y}_{l}$ as auxiliary inputs to guide the inpainting process. Besides, a binary mask $\mathbf{x}_{m}$ is used to represent missing pixels. Prior to training, the images in the dataset are preprocessed by combining them with masks to create the image with missing pixels $(\mathbf{x}_{r},\mathbf{x}_{d},\mathbf{x}_{e},\mathbf{x}_{l})$. 
Given the masked image samples as the input, the primary objective is to train a Hierarchical GAN that generates the recovered RGB images $\mathbf{r}_{r}$ and recovered depth images $\mathbf{r}_{d}$ with accurate inpainting of missing regions and as close as original RGB images $\mathbf{y}_{r}$ and depth images $\mathbf{y}_{d}$.

\subsection{Generative Adversarial Networks}
GAN comprises two essential components: a generator and a discriminator. Generator $G$ learns the underlying distribution of real data and produces the synthetic data samples $r$ by transforming condition information $\mathbf{x}$ into a meaningful data representation. Discriminator $D$ acts as a classifier, determining whether a given sample is from real data distribution $\mathbf{y}$ or generator. These two components engage in a min-max game:
\begin{gather}
    \min_G \max_D \mathbb{E}_\mathbf{y} \log(D(\mathbf{y})) + \mathbb{E}_{\mathbf{x}} \log(1-D(G(\mathbf{x}))
\end{gather}
The discriminator aims to maximize the probability of correctly assigning label to the input samples, while the generator seeks to minimize difference between synthetic and real sample to deceive the discriminator. The adversarial training continues until the synthetic sample become virtually indistinguishable from real data.

\section{Hierarchical Inpaiting GAN}

\subsection{Overview}

We present \emph{HI-GAN} for RGB and Depth image inpainting, an innovative framework that comprises three GANs collaborating together to recover missing pixels and construct target images. 

\emph{HI-GAN} is formulated and trained in an end-to-end manner, with masked RGB, depth, edge, and label segmentation images as input, and RGB and depth images with no missing pixels as output. 
Specifically, \emph{EdgeGAN} and \emph{LabelGAN} act as regularizers, focusing on inpainting edge images with missing pixels and semantic segmentation label images respectively; \emph{CombinedRGBD-GAN}, the primary model, inpaints RGB and Depth images with the assistance of regularizer models. It integrates edge and semantic segmentation information with filled missing pixels from \emph{EdgeGAN} and \emph{LabelGAN} to guide the recovering process and enhance the inpainting quality.

One of the key novelties lies in the hierarchical optimization process, which makes two auxiliary models, \emph{EdgeGAN} and \emph{LabelGAN}, more powerful regularizers. They are not only optimized during their individual training phases but also subjected to further optimization within \emph{CombinedRGBD-GAN} based on the RGBD inpainting performance. Compared with existing well-established models for image inpainting, \emph{HI-GAN} is designed to comprehensively learn the features from edge and label information and deliver robust inpainting outcomes.

\subsection{Network Architecture}
In this section, we describe the specific architecture inside regularizers and primary GAN that collaborate together for RGBD inpainting. Figure \ref{fig:hGAN} shows our proposed \emph{HI-GAN}.

\subsubsection{Regularizers: \emph{EdgeGAN} and \emph{LabelGAN}}
Figure \ref{fig:hGAN} presents the network architecture of \emph{EdgeGAN} and \emph{LabelGAN}. Both regularizers adhere to a typical GAN framework, consisting of a Generator and a Discriminator. 

Starting with \emph{EdgeGAN}, the generator $\mathcal{G}_{e}$ incorporates an encoder-decoder structure, aiming to produce inpainted edge images with filled missing pixels. Firstly, the masked edge image $\mathbf{x}_{e}$ serves as inputs. After flowing through gated convolution, gated residual blocks, instance normalization, and ReLU activations in the Edge encoder, the latent edge representation $\mathbf{z}_{e}$ is produced from Edge Encoder $E_{e}$. Subsequently, Edge Decoder $E_{e}$, exhibiting the same architecture as the encoding counterpart, learns $\mathbf{z}_{e}$ and reconstructs the inpainted edge images $\mathbf{r}_{e}$. The discriminator $\mathcal{D}_{e}$ is a classifier that estimates the likelihood $\mathbf{p}_{e}$ that a given sample is from the real dataset versus being inpainted from Edge Generator. It is structured as a patchGAN\cite{Isola} based discriminator with spectral normalization\cite{miyato2018spectral}, LeakyReLU activations, and a softmax layer. 

Parallel to \emph{EdgeGAN}, \emph{LabelGAN} shares the identical architecture paradigm, except for the input and output. In Label Generator $\mathcal{G}_{l}$, Label Encoder $E_{l}$ takes masked label images $\mathbf{x}_{l}$ as input and learns a latent label representation $\mathbf{z}_{l}$. It is then processed by Label Decoder $D_{l}$ to generate inpainted label segmentation image $\mathbf{r}_{l}$. The discriminator $\mathcal{D}_{l}$ performs as a classifier to differentiate between inpainted and original label images.

\emph{EdgeGAN} and \emph{LabelGAN} act as powerful regularizers, as their encoders provide crucial latent representations to \emph{CombinedRGBD-GAN} and guide the generation of high-quality inpainted RGB and depth images. This collaborative interplay will be introduced in the optimization process. 

\subsubsection{\emph{CombinedRGBD-GAN}}

\emph{CombinedRGBD-GAN} aims to inpaint masked RGB and depth images, with the assistance of \emph{EdgeGAN} and \emph{LabelGAN} encoder to boost the performance. It also follows a generator $\mathcal{G}_r$ with an encoder-decoder mechanism and discriminator $\mathcal{D}_r$. 
In the generator $\mathcal{G}_r$, RGB Encoder $E_{r}$ and Depth encoder $E_{d}$ take masked RGB images $\mathbf{x}_{r}$ and masked depth images $\mathbf{x}_{d}$ as input respectively, and produce a latent RGB representation $\mathbf{z}_{r}$ and a latent depth representation  $\mathbf{z}_{d}$ separately. Together with the representations $\mathbf{z}_{e}$ and $\mathbf{z}_{l}$ as the auxiliary input from encoders of \emph{EdgeGAN} and \emph{LabelGAN}, they are concatenated through Fusion module and fed into the final decoder $D_{r}$ and reconstruct the final RGB image $\mathbf{r}_r$ and depth image $\mathbf{r}_d$ with no missing pixels. 
Both encoders $E_{r}$ and $E_{d}$ in the generator $\mathcal{G}_{r}$ consist of gated convolutions, instance normalization, ReLU activations, along with gated residual blocks to extract complex and relevant information from the masked RGB and depth images. 
RGBD Decoder $D_{r}$ encompasses gated upsampling convolutions, instance normalization, and Relu activations. To distinguish between generated and original images, a PatchGAN-based discriminator $\mathcal{D}_{r}$ with spectral normalization and LeakyReLU activations serves as a classifier to output probability $\mathbf{p}_{r}$. 

\subsection{Hierarchical Optimization}
The optimization strategy of \emph{HI-GAN} performs in a hierarchical fashion. Due to the incorporation of latent edge and label representations in \emph{CombinedRGBD-GAN}, the encoders of \emph{EdgeGAN} and \emph{LabelGAN} are optimized further beyond their individual adversarial feedback. 

This approach offers two distinct advantages from two perspectives. (1) Inpainted edge and label images from generators of \emph{EdgeGAN} and \emph{LabelGAN} may exhibit subpar quality due to insufficient training. Directly utilizing them is less effective to augment RGBD inpainting. While, latent representations of \emph{EdgeGAN} and \emph{LabelGAN} encoders also extract relevant features from the edge and label images, and allow for optimization to improve their quality during the training process.
Therefore, it is reasonable and feasible to incorporate latent edge and label features into \emph{CombinedRGBD-GAN}.
(2) Individual optimizations of regularizers are based on the performance of inpainted auxiliary images only, while the integrated optimizations are based on the performance of our ultimate object, inpainted RGB and depth images. The dual optimization reinforces the roles of \emph{EdgeGAN} and \emph{LabelGAN} as influential regularizers and guide the generation of inpainted images in an end-to-end manner. 


\subsubsection{Individual Training on Regularizers}
During the individual training inside \emph{EdgeGAN} and \emph{LabelGAN}, the objective function for the regularizer encompasses Adversarial Loss and Feature Matching (FM) loss. The adversarial loss for generator and discriminator are calculated as follows:
\begin{gather}
    \mathcal L^e_{a,\mathcal{G}}= \mathbb{E}_{\mathbf{x}_e}[\log  \mathcal{D}_e( \mathcal{G}_e(\mathbf{x}_e))] \\
    \mathcal L^e_{a,\mathcal{D}}=\mathbb{E}_{\mathbf{y}_e}[\log  \mathcal{D}_e(\mathbf{y}_e)] + \mathbb{E}_{\mathbf{x}_e}[\log (1- \mathcal{D}_e( \mathcal{G}_e(\mathbf{x}_e)))] \\
    \mathcal L^l_{a,\mathcal{G}}= \mathbb{E}_{\mathbf{x}_l}[\log  \mathcal{D}_l( \mathcal{G}_l(\mathbf{x}_l))] \\
    \mathcal L^l_{a,\mathcal{D}}=\mathbb{E}_{\mathbf{y}_l}[\log  \mathcal{D}_l(\mathbf{y}_l)] + \mathbb{E}_{\mathbf{x}_l}[\log (1- \mathcal{D}_l( \mathcal{G}_l(\mathbf{x}_l)))] 
\end{gather}


FM loss compares the generated output and the ground truth image at a feature level, allowing the model to produce images perceptually aligned with the ground truth:
\begin{gather}
    \mathcal{L}^e_{fm} = \left\|\mathbb{E}_{\mathbf{y}_e}  [\mathcal{D}_e(\mathbf{y}_e)]-\mathbb{E}_{\mathbf{x}_e} [\mathcal{D}_{e}( \mathcal{G}_e(\mathbf{x}_e))]\right\|_2^2 \\
    \mathcal{L}^l_{fm} = \left\|\mathbb{E}_{\mathbf{y}_l}  [\mathcal{D}_l(\mathbf{y}_l)]-\mathbb{E}_{\mathbf{x}_l} [\mathcal{D}_l( \mathcal{G}_l(\mathbf{x}_l))]\right\|_2^2   
\end{gather}
where $\mathbf{y}_e$ and $\mathbf{y}_l$ represent the ground truth edge and label images, $\mathcal{G}_e(\mathbf{x}_e))$ and $\mathcal{G}_l(\mathbf{x}_l))$ are inpainted edge and label images with no missing pixels from the generators given the masked images $\mathbf{x}_e$ and $\mathbf{x}_l$, $\mathcal{D}_e(\mathbf{x_e})$ and $\mathcal{D}_l(\mathbf{x_l})$ denote probability from discriminators of two regularizers to distinguish the input source. $\lambda$ controls how much each loss term contributes to the overall loss value.

Therefore, the generators and discriminators of \emph{EdgeGAN} and \emph{LabelGAN} are presented as follows:
\begin{gather}
    \mathcal{L}^{e}_{\mathcal{G}} = \lambda^{e}_{a,\mathcal{G}} \mathcal{L}^{e}_{a,\mathcal{G}} + \lambda^{e}_{fm} \mathcal{L}^{e}_{fm} \\
    \mathcal L^{e}_\mathcal{D}= \lambda^{e}_{a,\mathcal{D}} \mathcal{L}^{e}_{a,\mathcal{D}}\\
    \mathcal{L}^{l}_{\mathcal{G}} = \lambda^{l}_{a,\mathcal{G}} \mathcal{L}^{l}_{a,\mathcal{G}} + \lambda^{l}_{fm} \mathcal{L}^{l}_{fm} \\
    \mathcal L^{l}_\mathcal{D}= \lambda^{l}_{a,\mathcal{D}}\mathcal{L}^{l}_{a,\mathcal{D}}
\end{gather}

\subsubsection{Incorporated Training on \emph{CombinedRGBD-GAN}}
For the training of \emph{CombinedRGBD-GAN}, the objective function consists of adversarial loss, FM loss, perceptual loss, and style loss. 

The adversarial loss and FM loss from recovered RGB and depth image is as follows:
\begin{gather}
    \mathcal L^r_{a,\mathcal{G}}= \mathbb{E}_{\mathbf{x}_r}[\log  \mathcal{D}_r( \mathcal{G}_r(\mathbf{x}_r))] \\
    \mathcal L^r_{a,\mathcal{D}}=\mathbb{E}_{\mathbf{y}_r}[\log  \mathcal{D}_r(\mathbf{y}_r)] + \mathbb{E}_{\mathbf{x}_r}[\log (1- \mathcal{D}_r( \mathcal{G}(\mathbf{x}_r)))] \\
    \mathcal L^d_{a,\mathcal{G}}= \mathbb{E}_{\mathbf{x}_d}[\log  \mathcal{D}_r( \mathcal{G}_r(\mathbf{x}_d))] \\
    \mathcal L^d_{a,\mathcal{D}}=\mathbb{E}_{\mathbf{y}_d}[\log  \mathcal{D}_r(\mathbf{y}_d)] + \mathbb{E}_{\mathbf{x}_d}[\log (1- \mathcal{D}_r( \mathcal{G}_r(\mathbf{x}_d)))] \\
    \mathcal{L}^r_{fm} = \left\|\mathbb{E}_{\mathbf{y}_r}  [\mathcal{D}_r(\mathbf{y}_r)]-\mathbb{E}_{\mathbf{x}_r} [\mathcal{D}_{r}( \mathcal{G}_r(\mathbf{x}_r))]\right\|_2^2 \\
    \mathcal{L}^d_{fm} = \left\|\mathbb{E}_{\mathbf{y}_d}  [\mathcal{D}_r(\mathbf{y}_d)]-\mathbb{E}_{\mathbf{x}_d} [\mathcal{D}_r( \mathcal{G}_r(\mathbf{x}_d))]\right\|_2^2   
\end{gather}

Perceptual loss or VGG loss \cite{PerceptualLoss} uses pre-trained VGG-16\cite{VGG16Simonyan15} model with pre-trained ImageNet weights to extract relevant higher-level feature representations from the predicted image  and ground truth image at different layers and calculates their $L_1$ distance. This ensures that the filled image and the ground truth image have similar feature representations. The perceptual loss for RGB and depth images are given as: 
\begin{gather}
\begin{aligned}
    \mathcal{L}^r_{p} &=\sum_{n=0}^{N-1}\|\Psi_n^{\mathbf{r}_r}-\Psi_n^{\mathbf{y}_r}\|_1 \\
    \mathcal{L}^d_{p} &=\sum_{n=0}^{N-1}\|\Psi_n^{\mathbf{r}_d}-\Psi_n^{\mathbf{y}_d}\|_1 \\ 
\end{aligned}
\end{gather}
where $\Psi_n$ is the feature maps of the N$^{th}$ selected layer computed by pre-trained VGG-16. 

Style Loss\cite{gatys2015neural} is computed by feature maps given by a pre-trained VGG-16 model. The Gram Matrix is computed for each feature map for the predicted and ground truth image, respectively, before computing the Frobenius norm:

\begin{gather}
\mathcal{L}^r_{s}=\sum_{n=0}^{N-1} \| K_n((\Psi_n^{\mathbf{r}_r})^{\top}(\Psi_n^{\mathbf{r}_r}) -(\Psi_n^{\mathbf{y}_{r}})^{\top}(\Psi_n^{\mathbf{y}_{r}})) \|_1 \\
\mathcal{L}^d_{s}=\sum_{n=0}^{N-1} \| K_n((\Psi_n^{\mathbf{r}_d})^{\top}(\Psi_n^{\mathbf{r}_d}) -(\Psi_n^{\mathbf{y}_{d}})^{\top}(\Psi_n^{\mathbf{y}_{d}})) \|_1
\end{gather}
where $\Psi_n$ represents high level features and $K_n$ is the normalization factor for the $N^{th}$ selected layer.

Thus, the objective function for \emph{CombinedRGBD-GAN} of generator $\mathcal{L}^c_{\mathcal{G}}$ contains the feedback from RGB image $\mathcal{L}^{r}_{\mathcal{G}}$ and depth image $ \mathcal{L}^{d}_{\mathcal{G}}$, similar for that of discriminator $\mathcal L^{c}_\mathcal{D}$:
\begin{gather}
    \mathcal{L}^{r}_{\mathcal{G}} = \lambda^{r}_{a,\mathcal{G}} \mathcal{L}^{r}_{a,\mathcal{G}} + \lambda^{r}_{fm} \mathcal{L}^{r}_{fm} + \lambda^{r}_{p} \mathcal{L}^{r}_{p} + \lambda^{r}_{s} \mathcal{L}^{r}_{s} \\
    \mathcal{L}^{d}_{\mathcal{G}} = \lambda^{d}_{a,\mathcal{G}} \mathcal{L}^{d}_{a,\mathcal{G}} + \lambda^{d}_{fm} \mathcal{L}^{d}_{fm} + \lambda^{d}_{p} \mathcal{L}^{d}_{p} + \lambda^{d}_{s} \mathcal{L}^{d}_{s} \\
    \mathcal{L}^c_{\mathcal{G}} = \mathcal{L}^{r}_{\mathcal{G}} +\mathcal{L}^{d}_{\mathcal{G}} \\
    \mathcal L^{c}_\mathcal{D}= \lambda^{r}_{a,\mathcal{D}} \mathcal{L}^{r}_{a,\mathcal{D}} + \lambda^{d}_{a,\mathcal{D}} \mathcal{L}^{d}_{a,\mathcal{D}}
\end{gather}


\section{Evaluation}
\begin{figure*}[h]
    \centering
    \includegraphics[width=1.0\textwidth]{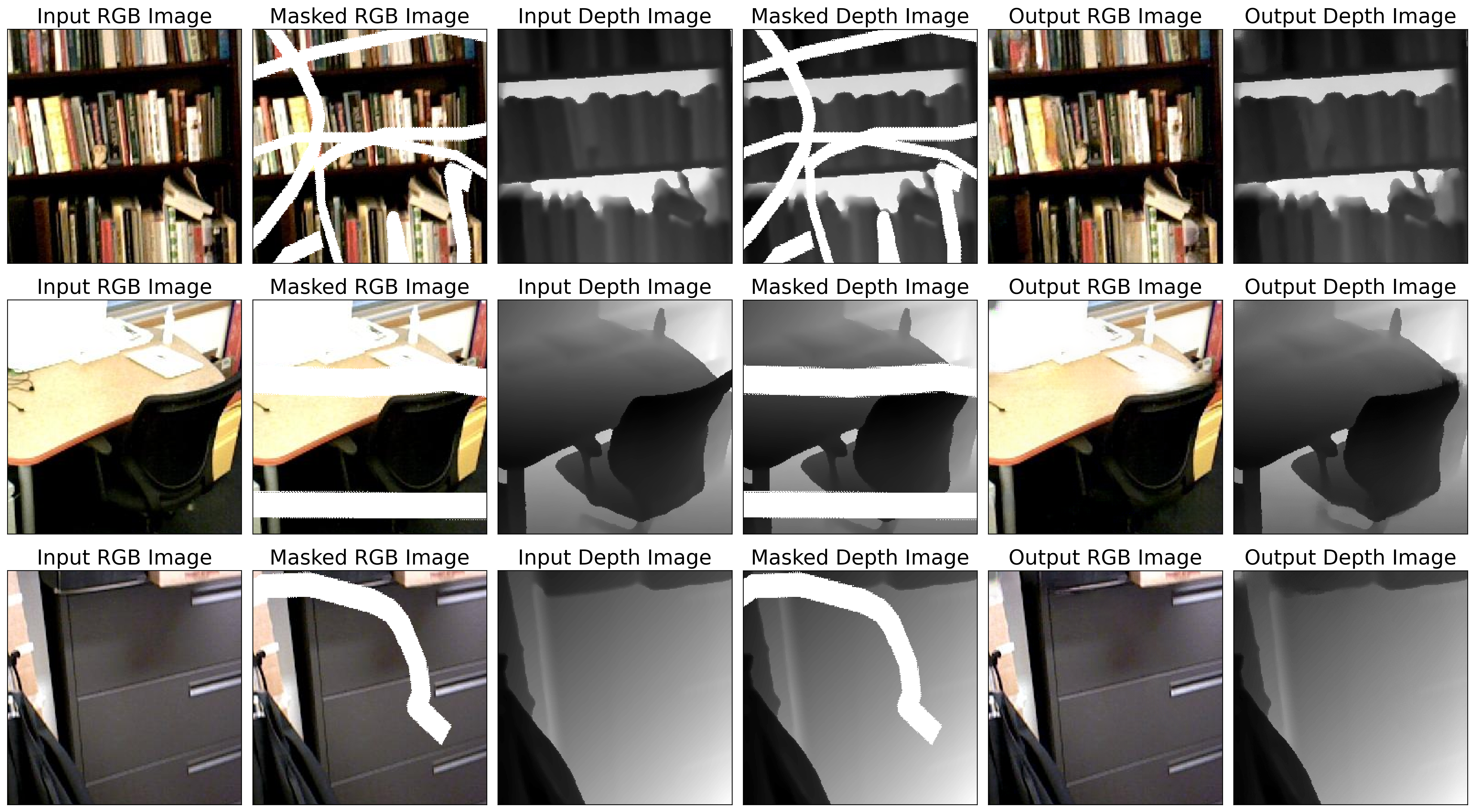}
    \caption{\emph{HI-GAN} results for RGBD inpainting. Each row is a sample randomly selected from the test dataset. }
    \label{fig:hgan REs}
\end{figure*}

\subsection{Experiment Setup}
\subsubsection{Dataset} We use the SUN-RGBD \cite{SilbermanECCV12,sunRGBD,B3DO,6751312} dataset for model evaluation, including 9500 images for training and validation, as well as 835 images for testing. The SunRGBD dataset contains RGB images along with Depth and Semantic segmentation labels. To generate edge images, we adopt the Canny edge detector with the threshold values set to 100 and 200 respectively. The lower threshold value of 100 and the upper threshold value of 200 gives us the best edge images and best model performance. To generate mask images, we leverage the QD-IMD, Quick Draw Irregular Mask Dataset \cite{QD-IMD}, containing irregular masks of varying complexity and based on the Quick Draw dataset. During data preprocessing, RGB images are sized to dimensions of $256\times256\times3$; Depth, Edge, Label, and Mask images are resized to $256\times256$.

\subsubsection{Baselines}
To compare our results with other state-of-the-art methods, the following baselines are chosen: (1)
\textbf{Partial Convolution (PConv)} \cite{liu2018partialinpainting}: an image inpainting technique using convolutional neural networks to accurately fill in missing regions by selectively updating image parts and incorporating masked regions for seamless restoration.
(2) \textbf{EdgeConnect} \cite{Nazeri_2019_ICCV}: With a two-step process, EdgeConnect predicts the missing edges and generates the inpainted regions based on the input and edge maps.

\begin{table*}[ht]
\begin{center}
\caption{Numerical comparison of \emph{HI-GAN} and baselines on test dataset.}
\begin{tabular}{@{}lllllllll@{}}
\hline
Model & RGB SSIM  & RGB PSNR &  Depth PSNR & RGB MAE & RGB RMSE & Depth MAE & Depth RMSE  \\
\hline
PConv     &  0.8651  & 27.5033    & N/A & 7.6075 & 11.4138  &  N/A & N/A  \\
EdgeConnect    &   0.5264   &  14.2957     &  N/A & 36.8311  & 53.9073  &  N/A & N/A \\
\emph{HI-GAN}   &  \textbf{0.9476}  & \textbf{30.38}   & 34.257 & \textbf{5.3687} & \textbf{8.901} & 2.5059  & 5.582\\
\hline
\end{tabular}
\label{tab:results}
\end{center}
\end{table*}

\subsubsection{Evaluation Metrics}
\emph{HI-GAN} and baselines are evaluated by following image evaluation metrics and traditional metrics:
(1) \textbf{Structural Similarity Index Measure (SSIM)} quantifies the similarity between two images. It measures both the structural information and the luminance similarity, taking into account human perception. The range of SSIM is $[-1,1]$, where 1 indicates a perfect match, while lower values indicate increasing dissimilarity. 
(2) \textbf{Peak Signal to Noise Ratio (PSNR)} is a widely used metric in image and video processing to assess the quality of a reconstructed or compressed image or video compared to its original version. It is measured in decibels (dB). Higher PSNR values imply greater quality and a closer resemblance between the original and processed images. Conversely, lower values indicate more significant distortion or information loss.
(3) \textbf{Mean Absolute Error (MAE)} calculates the average absolute pixel-wise difference between generated and ground truth images.
(4) \textbf{Root Mean Squared Error (RMSE)} calculates the square root of the average of squared differences between prediction and actual observation.

\begin{figure*}[h!]
    \centering
    \includegraphics[width=1.0\textwidth]{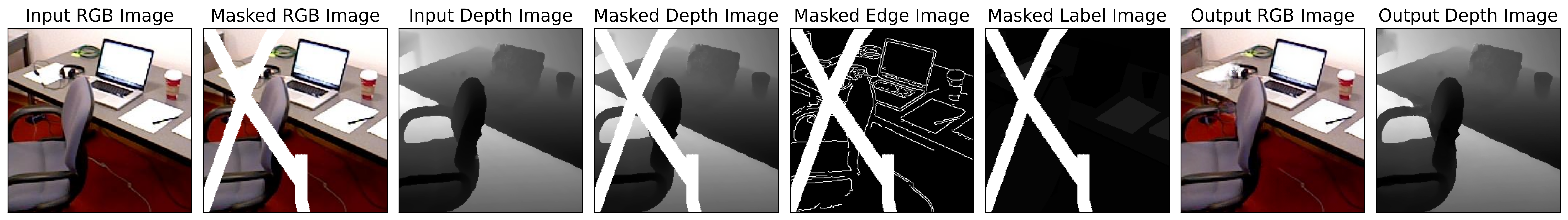}
    \caption{\emph{HI-GAN} results for RGBD inpainting with all auxiliary inputs. }
    \label{fig:hgan REs all inputs}
\end{figure*}

\subsubsection{Implementation Details}
\emph{HI-GAN} is trained for 500,000 epochs, using 4 RTX Titan GPUs with 24 GB of VRAM. 
The batch size is set to 2. We choose Adam optimizer on all three components (\emph{EdgeGAN}, \emph{LabelGAN}, \emph{CombinedRGBD-GAN}) with a learning rate of 0.0001.
For the weight of FM loss, 
\emph{EdgeGAN} is 10.0, \emph{LabelGAN} is 15.0, RGB in \emph{CombinedRGBD-GAN} is set to 5.0, and depth is set to 5.0.
For perceptual loss, RGB is set to 3.0, and depth is set to 2.0. For the weight of style loss in \emph{CombinedRGBD-GAN}, RGB is set to 2.0, and depth is 3.0. For adversarial loss weight, RGB is set to 0.005, and depth is set to 0.01. The baselines are also trained following their default parameters and evaluated on the test set.

\subsection{Results and Discussion}
In this section, we showcase the visualized inpainting images, compare the results with baselines, and provide an in-depth analysis of model components.
\subsubsection{Visualization of Inpainting images.}
\begin{figure}[h!]
    \centering
    \includegraphics[width=0.5\textwidth]{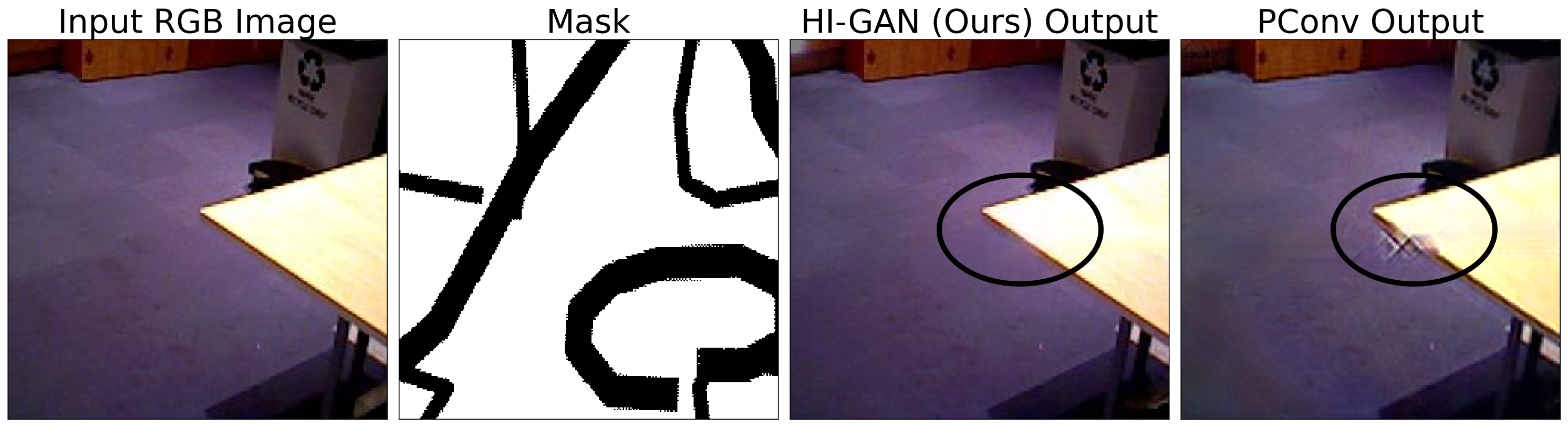}
    \caption{\emph{HI-GAN} and PConv output comparison. The black circle represents the region of interest, showing the superiority of \emph{HI-GAN} over PConv.}
    \label{fig:allComparison}
\end{figure}

Figure \ref{fig:hgan REs} presents the inpainted images from \emph{HI-GAN}. Each row represents an example of distinct pairs of RGB and Depth images. \emph{HI-GAN} successfully completes the missing pixel regions, generating output images seamlessly aligned with the ground truth counterparts. This visual evidence demonstrates that even without any costly attention mechanism, our model can still capture local and global information and preserve the edge and semantic information while filling in the missing pixels. This achievement is attributed to the incorporation of edge and label images. These auxiliary inputs not only provide relevant insights about boundary and object segmentation categories, but also serves to guide the joint RGB and Depth inpainting during optimization. 
Figure \ref{fig:hgan REs all inputs} shows all of the inputs including auxiliary inputs, such as masked edge and label images, and outputs from our model. 
Figure \ref{fig:allComparison} demonstrates the effectiveness of \emph{HI-GAN} in preserving boundary and object-specific information, highlighted within the black circle. \emph{HI-GAN} can capture boundary and object context information using edge and label information to provide improved inpainting results. More images for comparison are in Supplementary.

\subsubsection{Numerical Comparison.}
Table \ref{tab:results} quantifies the performance of \emph{HI-GAN} and baselines, we have the following observations. (1) Our model significantly outperforms the cutting-edge models regarding SSIM and PSNR. We perform the analysis solely on RGB outputs, because PConv and EdgeConnect generate RGB images only. This is one of the issues with many state-of-the-art image inpainting techniques as we have discussed. (2) PConv exhibits a reasonable well performance, with SSIM and PSNR values of 0.8651 and 27.5033. While EdgeConnect performs poorly. It reveals that EdgeConnect is very sensitive to the separate training of edge and label models, as well as their subsequent integration into the Inpainting model's training. Even extended training epochs on two baselines cannot yield improvements.
(3) \emph{HI-GAN} achieves an MAE of 5.3687 and an RMSE of 8.901, surpassing the chosen baselines. These lower error values underscore the model's precision and superior performance. (4) Moreover, \emph{HI-GAN} can accurately inpaint depth images with an impressive low MAE of 2.5059 and RMSE of 5.582. While PConv performs reasonably well, EdgeConnect shortcomings persist due to its intricate and demanding training process, which is consistent with the performance of SSIM and PSNR.


\subsubsection{Ablation Study}
\emph{EdgeGAN} and \emph{LabelGAN} are designed to improve the overall RGBD inpainting. We conduct experiments on two variants of \emph{HI-GAN} to demonstrate their effectiveness: \textbf{\emph{HI-GAN}-E} is trained without \emph{LabelGAN} and \textbf{\emph{HI-GAN}-L} is the variant without \emph{EdgeGAN}. Table \ref{tab:abalation}. shows the quantitative comparison. 
(1) It reveals that \emph{EdgeGAN} and \emph{LabelGAN} boost the performance of RGBD inpainting. With the help of \emph{EdgeGAN} and \emph{LabelGAN}, \emph{HI-GAN} achieves the highest SSIM and  PSNR for inpainted RGB images. 
(2) \emph{HI-GAN}-E maintains high SSIM and PSNR for recovered RGB and Depth images, followed by \emph{HI-GAN}-L. This observation demonstrates the potency of semantic label images as influential auxiliary inputs, capable of fortifying the RGBD image inpainting process. Since depth images also inherently provide boundary information, \emph{HI-GAN}-L relying solely on edge images as auxiliary input fall behind in performance when compared with the others. 
Our work proves that semantic segmentation label image is a worthy candidate as auxiliary inputs, capable of augmenting both RGB and Depth inpainting.  The simultaneous incorporation of label and edge images, combined with our hierarchical training approach, culminates in superior inpainting performance, achieved without any attention mechanism.

\begin{table}[h!]
\begin{center}
\caption{Numerical results on ablation study}\label{tab1}%
\begin{tabular}{@{}lllllll@{}}
\hline
Model & RGB SSIM  & RGB PSNR  & Depth PSNR  \\
\hline
\emph{HI-GAN}    &   \textbf{0.9461}  & \textbf{30.38}   & \textbf{34.257}  \\
\emph{HI-GAN}-L    &  0.9422  &  28.9947   & 34.15 \\
\emph{HI-GAN}-E   &  0.9204  &  26.891   & 31.4472 \\
\hline
\end{tabular}
\label{tab:abalation}
\end{center}
\end{table}

\section{Conclusion}
We propose \emph{HI-GAN}, a composite framework with three GANs working cohesively in an end-to-end manner for RGBD inpainting. The integration of edge images and especially label images improves context comprehension through boundary delineation, specific object information, and object-aware inpainting. Moreover, the hierarchical optimization of \emph{EdgeGAN} and \emph{LabelGAN} inside \emph{CombinedRGBD-GAN} guides the inpainting procedure. 
\emph{HI-GAN} achieves the highest SSIM and PSNR, as well as the lowest MAE and RMSE, outperforming baseline models, without any attention mechanisms. Ablation study proves the efficacy of edge and label images along with \emph{EdgeGAN} and \emph{LabelGAN} in RGBD inpainting. 
Our future work will incorporate attention mechanisms and evaluate the performance against other attention-based baselines for RGB and Depth inpainting scenarios.

\bibliography{references}

\end{document}